\documentclass[runningheads]{llncs}
\usepackage{graphicx}
\usepackage{bm}
\usepackage{multicol}
\usepackage{amsmath, amssymb}
\usepackage{graphicx}
\usepackage{subfigure}
\usepackage{caption}
\usepackage[numbers,sort]{natbib}

\begin{document}
\title{Extending Temporal Data Augmentation for Video Action Recognition}
%
%
\author{Artjoms Gorpincenko\orcidID{0000-0001-7853-8458} \\ \and
Michal Mackiewicz\orcidID{0000-0002-8777-8880}}
\authorrunning{A. Gorpincenko and M. Mackiewicz}
%
\institute{School of Computing Sciences, \\University of East Anglia, \\Norwich, \\England.\\
\email{\{a.gorpincenko,m.mackiewicz\}@uea.ac.uk}}
\maketitle              
\begin{abstract}
Pixel space augmentation has grown in popularity in many Deep Learning areas, due to its effectiveness, simplicity, and low computational cost. Data augmentation for videos, however, still remains an under-explored research topic, as most works have been treating inputs as stacks of static images rather than temporally linked series of data. Recently, it has been shown that involving the time dimension when designing augmentations can be superior to its spatial-only variants for video action recognition \cite{temporal}. In this paper, we propose several novel enhancements to these techniques to strengthen the relationship between the spatial and temporal domains and achieve a deeper level of perturbations. The video action recognition results of our techniques outperform their respective variants in Top-1 and Top-5 settings on the UCF-101 \cite{ucf} and the HMDB-51 \cite{hmdb} datasets.
\keywords{Data augmentation  \and Temporal domain \and Action recognition.}
\end{abstract}
\section{\uppercase{Introduction}}\label{introduction}
Deep convolution neural networks (CNNs) have become the standard approach for a large number of computer vision tasks, by virtue of their unique ability to learn the most useful features from the data in the unmanned manner. However, large amounts of diverse labeled training imagery are usually required to guarantee models' high accuracy, which are often unavailable. Acquiring and annotating new data is generally expensive, time-consuming, and sometimes even impossible, resulting in networks underfitting or overfitting, depending on the training set variance. In recent years, several deep learning areas have been explored to tackle the aforementioned problems, such as domain adaptation \cite{dann,dirt,se,vmt,adda}, network regularization \cite{dropout,earlystopping,weightdecay,simpleweightdecay}, data generation \cite{dagan,lowgan,medgan,jellygan}, and data augmentation \cite{randaugment,vat,cutmix,cutout,mixup}, all showing significant performance gains over their respective baselines. 

Due to its ability of expanding and populating the training distribution through synthetically created samples, pixel space augmentation was successfully used as the main driver in a number of semi-supervised \cite{mixup,cutmix,vat,fixmatch}, self-supervised \cite{contr,momentum,self}, and domain adaptation \cite{se,dirt,vmt} studies. The use of feature space augmentation was also explored for both static and sequential imagery \cite{feataugment,datasetaugment,feattransfer,vatta3n}, yielding improvements in models' accuracy. Data augmentation for videos, however, still remains an under-explored research area, as most works have been treating inputs as stacks of static images rather than temporally linked series of data. A recent study has shown that the time domain consideration while designing augmentations can be superior to its spatial-only variants for video recognition \cite{temporal}. 

In this paper, we expand on the previous work \cite{temporal}. We argue that some of the proposed techniques can be extended even further to fully utilise the time domain and achieve a deeper level of temporal perturbations, which results in more accurate and robust classifiers. The contributions of this paper can be summarised as follows:
\begin{enumerate}
    \item We expand the list of available augmentations in RandAugment-T \cite{temporal} by adding VideoReverse, FrameFadeIn, and VideoCutMix, augmentations that are video-specific and are done within a single sample;
    \item We increase the amount of magnitude checkpoints for all augmentation techniques to allow for non-linear temporal perturbations;
    \item We propose to linearly change the bounding pox positions for cut-and-paste algorithms, such as CutOut \cite{cutout}, CutMix \cite{cutmix}, and CutMixUp \cite{cutmixup}, and their extensions, as well as the mixing ratio in MixUp \cite{mixup} and CutMixUp \cite{cutmixup} extensions;
    \item The recognition results of the aforementioned techniques on the UCF-101 \cite{ucf} and the HMDB-51 \cite{hmdb} datasets either maintain competitive or exceed performance achieved by the previous work \cite{temporal}.
\end{enumerate}

\section{\uppercase{Related work}}\label{related_work}
\subsection{Spatial augmentation}
The earliest experiments that demonstrate the effectiveness of data augmentation are based on basic image modifications, such as axis flipping, rotations, translations, random cropping, and colour space alterations \cite{imagenet,multicolumn,visualcontana,visualobjcls}. These techniques are easy to implement, bear minimum computational overhead and are very likely to preserve the label after transformation. However, combining the aforementioned operations together can result in heavily inflated datasets and high risk of label warp. Therefore, a number of studies has been done on search algorithms that aim to find the optimal subset of augmentations for a particular task \cite{datasetaugment,smartaugment,bayesian,autoaugment}. Finally, RandAugment \cite{randaugment} presents an efficient framework that works out of the box for applying operations sequentially and without a separate search phase.

Image mixing is an approach that involves blending a pair of samples into one, enforcing the classifier to behave linearly in-between training data points. Performance gains can be observed even by averaging pixel values of two random images and retaining only one out of the two labels \cite{pairing}. This idea was further extended to more sophisticated techniques which proposed mixing at different ratios and working with soft labels \cite{mixup,cutmix,cowmix}, as well as their non-linear derivatives \cite{improvedmix}.

Adding small amounts of noise to the input images during training encourages CNNs to have smoother and stronger decision boundaries on the data manifold and results in learning more robust features \cite{noise}. The concept was thoroughly studied in the field of adversarial attacks, where the rival network's objective is to learn augmentations that result in misclassifications in the classification model \cite{deepfool,adversarialexamples,vat}. 

Creating synthetic data with the help of generative adversarial networks (GANs) \cite{gan} is yet another way to augment a dataset. With recent advancements in the field, GANs are now able to generate images that look real to human observers, in spite of illustrating entities that are not present in the training set \cite{stylegan,betterstylegan,largegan,jellygan}. The GAN framework also can be extended to improve the quality of samples created by variational auto-encoders \cite{autoencoders} or perform style transfer to map existing imagery to the domain of interest \cite{cyclegan,sonarfirst,sonarsecond}.

\subsection{Video recognition}
A clear-cut approach to video classification using CNNs is to include the temporal domain by extending the dimensionality of convolutional operations. 3D filters achieved superior results when compared to 2D, proving that the time domain has a lot of value \cite{3dcnn}. The inclusion of the temporal axis opened up a whole research area that is aimed at exploring its various fusion techniques. The most popular ones are slow fusion to improve the time awareness of the model \cite{slowfusion}, late fusion, where temporal features are blended at the last layer \cite{slowfusion}, longer fusion, which explores the benefit of extending the temporal depth \cite{longfusion}, and ensembling networks with different temporal awareness \cite{longfusion}. Finally, a combination of 2D an 1D kernels is proposed to substantially reduce the amount of learnable parameters without any loss in performance \cite{2d1d}.

Motivated by the fact that humans use different streams to process appearance and motion data, multiple stream models were proposed \cite{twostream}. The aim is to have separate spatial and temporal tracks, hence making it easier to encode relevant features in the respective streams. This is further enhanced by supplying different inputs - whereas the spatial path takes RGB frames, which contain appearance information, the temporal path receives optical flow frames that contain motion data. Later work shows that earlier fusion of the streams allows to retain the performance while halving the amount of learnable parameters \cite{twostreamfusion}.

\subsection{Temporal augmentation}
Although a substantial amount of work has been done on spatial augmentation, the field of temporal augmentation remains under-explored. Random Mean Scaling \cite{rms} stochastically varies the low-frequency feature components to regularize classifiers, whereas FreqAug \cite{spatiotemporal} experiments with randomly removing them. RandAugment-T \cite{temporal} extends the spatial-only framework to the time dimension and presents a set of modifications on cut-and-paste and blend algorithms, such as CutOut \cite{cutout}, CutMix \cite{cutmix}, MixUp \cite{mixup}, and CutMixUp \cite{cutmixup}, to produce temporally localisable features. Our work expands on the latter and proposes a set of modifications that can be used to make video classifiers more robust and accurate.

\section{\uppercase{Methods}}\label{method}
\subsection{Single video augmentation}
RandAugment \cite{randaugment} is an automated data augmentation framework that randomly selects a number of transformations for a given image. From a list of $K$ operations, RandAugment takes $N$ augmentations with the magnitude of $M$. Each transformation has a probability of $\frac{1}{K}$ to be chosen. A total of $K=14$ operations are presented: Identity, Rotate, Posterise, Equalise, Sharpness, Translate-X, Translate-Y, Colour, AutoContrast, Solarise, Contrast, Brightness, Shear-X, and Shear-Y.

RandAugment-T \cite{temporal} introduces $M_1$ and $M_n$, two magnitude points that are placed at the start and the end of each video. This allows for smooth augmentation transitions across the frames and brings the temporal component to the equation, where possible. The work also extends the list of available transformations by including ColourInvert, albeit it having static magnitude. All the operations mentioned above are taken directly from image augmentation, and are applied to a single video. Operations such as Identity, AutoContrast, Equalise, and ColourInvert do not have varying $M$, and hence are applied evenly across the sample. 

\begin{figure}
\includegraphics[width=\linewidth]{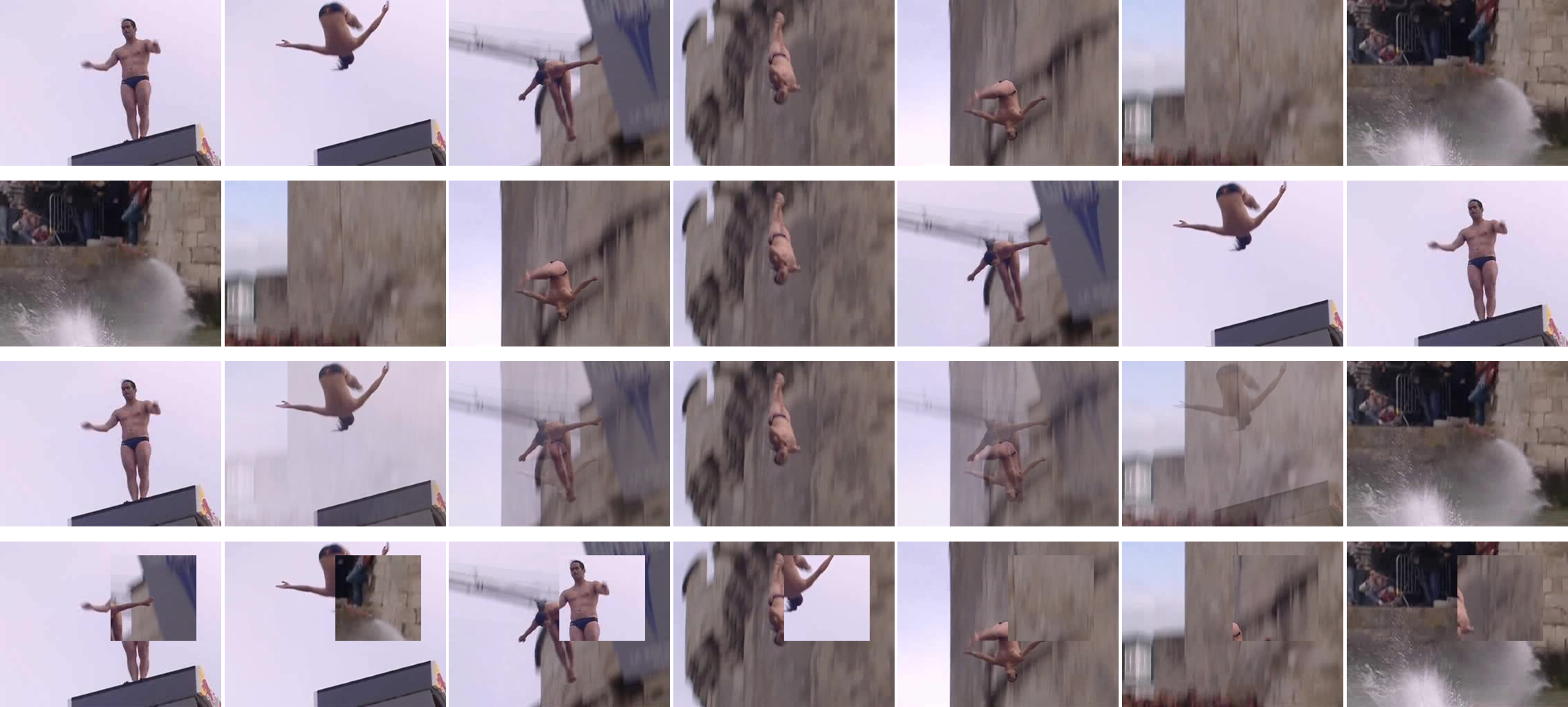}
   \caption{Unaugmented video, VideoReverse, FrameFadeIn, and VideoCutMix in the 1\textsuperscript{st}, 2\textsuperscript{nd}, 3\textsuperscript{rd}, and 4\textsuperscript{th} row, respectively.}
\label{fig:singlevideoaugment}
\end{figure}

Although previous work sticks to the aforementioned list of transformations \cite{autoaugment,randaugment,population,fastautoaugment}, the purpose of this paper is to propose temporal augmentations, rather than suggest a new augmentation policy. Therefore, we expand the list of available operations by introducing VideoReverse, FrameFadeIn, and VideoCutMix (Fig. \ref{fig:singlevideoaugment}) -  transformations that are designed specifically for video samples. VideoReverse turns the video backwards, creating a rewind effect, yet maintaining the semantics and integrity of the sample. FrameFadeIn is inspired by FadeMixUp \cite{temporal}, with the main difference being the use of a single sample and a simpler mixing ratio calculation:
\begin{equation}
    \tilde{x_t} = (1 - \lambda_t) x_t + \lambda_t x_{n-t},
\end{equation}
where $\tilde{x}$, $x$, $n$, and $\lambda$ indicate the mixed data, original data, total number of frames, and mixing ratio, respectively. Unlike FadeMixUp, we do not sample start and end points for
$\lambda$ interpolation. Instead, we gradually increase it from 0 to 0.5 until the middle of the video, then decrease it back to 0:
\begin{equation}
    \lambda_t =
    \begin{cases}
    \frac{t}{n}, & \text{if } t\leq \frac{n}{2},\\
    \frac{n-t}{n}, & \text{otherwise.}
    \end{cases}
\end{equation}
This maintains a healthy trade-off between spatial and temporal perturbations - when the distance between frames is large, the mixing ratio is small, and vice versa. Although it is possible to use sampled magnitudes instead, it significantly increases the risk of breaking temporal consistency. VideoCutMix is a temporal extension of CutMix \cite{cutmix,temporal} that can be applied to a single video:
\begin{equation}
    \tilde{x_t} = M \odot x_t + (1 - M) \odot \hat{x_t},
\end{equation}
where $M$, $\hat{x}$, and $\odot$ denote the binary region mask indicating where to drop out or fill in from two separate frames, video with randomly shuffled frames, and element-wise multiplication, respectively. Although cut-and-pasting happens within the same sample, the nondeterministic nature introduces a certain risk of altering data to the point where semantics may be significantly damaged or lost. To keep it at minimum, we set the region ratio to $0.2$ of the original frame size and keep the position of the bounding box static. As with all single sample augmentations, labels remain unchanged in VideoReverse, FrameFadeIn, and VideoCutMix.

\subsection{MagAugment}
\begin{figure}
\centering  
\subfigure{\includegraphics[width=0.496\textwidth]{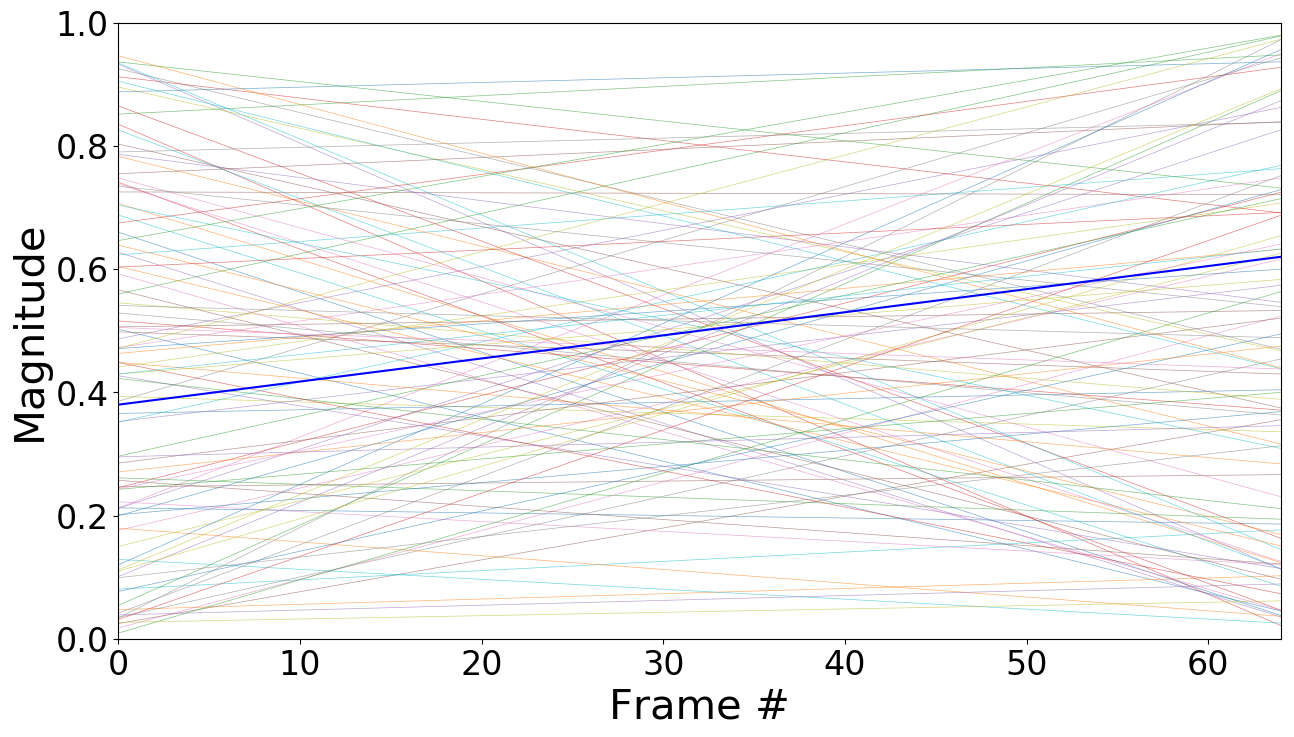}}
\subfigure{\includegraphics[width=0.496\textwidth]{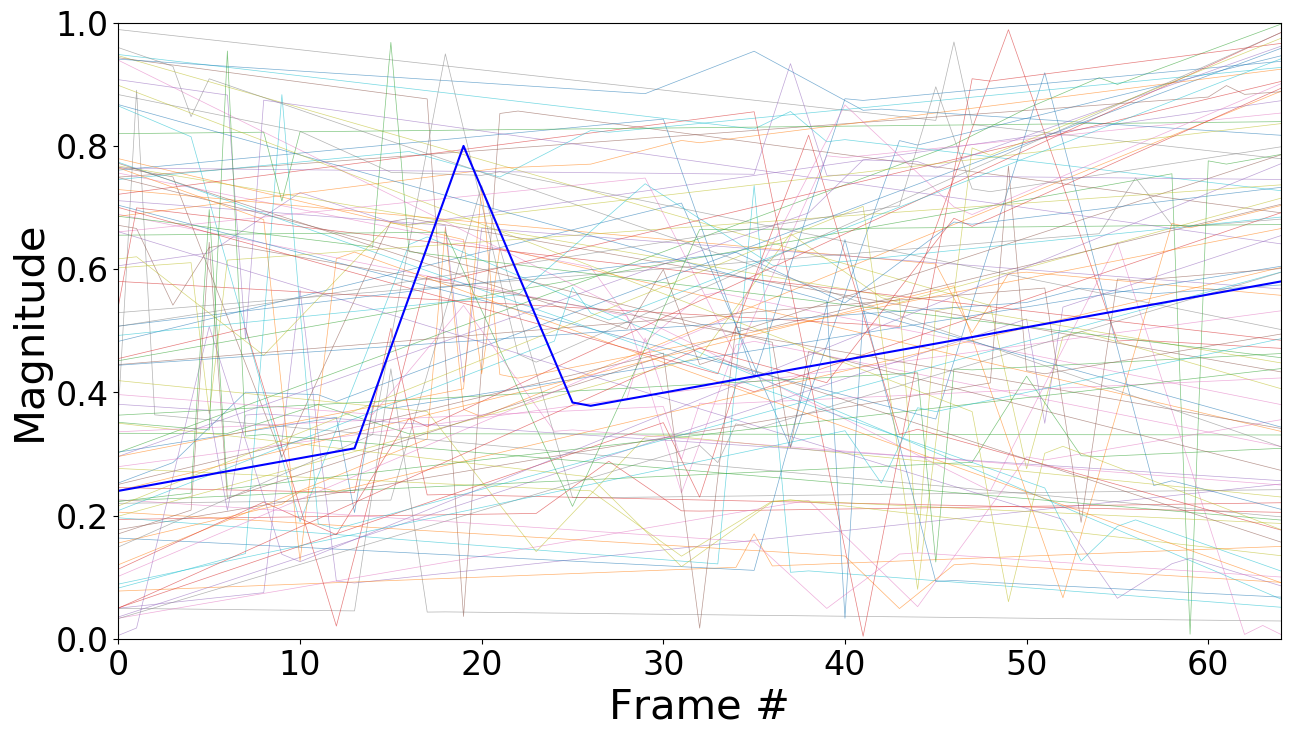}}
\caption{Visual comparison of magnitude manipulations proposed in RandAugment-T \cite{temporal} (left) and MagAugment (right), comprising of 100 randomly sampled $M$ arrays. Blue lines - randomly highlighted samples. For MagAugment, $\beta$ was set to 8.}
\label{fig:magaugment}
\end{figure}
RandAugment-T \cite{temporal} implements augmentation transitions across frames by putting two magnitude checkpoints, $M_1$ and $M_n$, at the start and the end of samples, and calculating the other $M_t$ via linear interpolation. The introduced change in magnitude leads to better video action recognition performances, when compared to its static variant \cite{temporal}. Our hypothesis is that having more magnitude checkpoints placed along the sample results in greater generalisation performance, as they are more likely to mimic perturbations observed in real-life conditions. Phenomena such as flashes, sudden camera shaking and/or movement, loss of focus, and exposure adjustments tend to happen in much shorter time periods than the length of the entire video. In this subsection, we propose MagAugment (Fig. \ref{fig:magaugment}) - a framework designed to increase the magnitude diversity even further, without interrupting the temporal consistency. 

We start with the linear signal connecting the two ends of the magnitude array. To introduce short and sporadic magnitude swings, we sample a point from the uniform distribution, $M_p \sim U(M_{min}, M_{max})$, where the parameters represent the minimum and maximum magnitude values for a given transformation. The duration of the perturbations in frames is set to $j \sim U(1, \beta)$, where $\beta$ is the MagAugment parameter. Finally, the location of the point is drawn from $p \sim U(1+j, n-j)$, where $n$ is the total amount of frames. The process can be repeated to model several fluctuations. To incorporate the magnitude swings into the original signal, we linearly interpolate from $M_{p-j}$ to $M_p$, then back to $M_{p+j}$. As a result, the overall augmentation direction is maintained, while allowing for occasional, more aggressive changes in pixel space that do not necessarily follow the general trend. 

\subsection{Temporal deleting, cut-and-pasting, and blending}
\setcounter{figure}{2}
\begin{figure}
\ContinuedFloat
\centering  
\subfigure[Top: CutOut \cite{cutout,temporal}, Bottom: FloatCutOut]{\includegraphics[width=\textwidth]{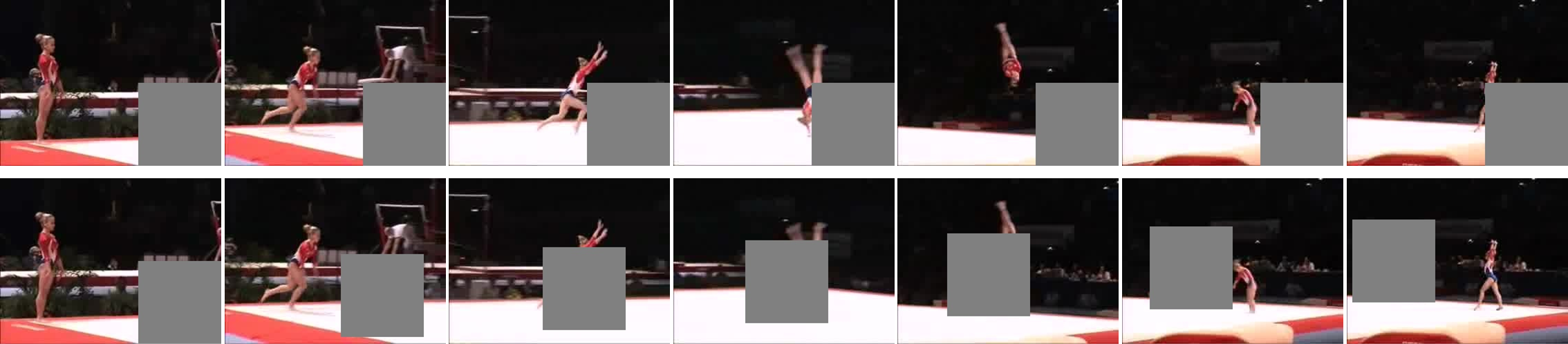}}
\subfigure[Top: CubeCutOut \cite{temporal}, Bottom: FloatCubeCutOut]{\includegraphics[width=\textwidth]{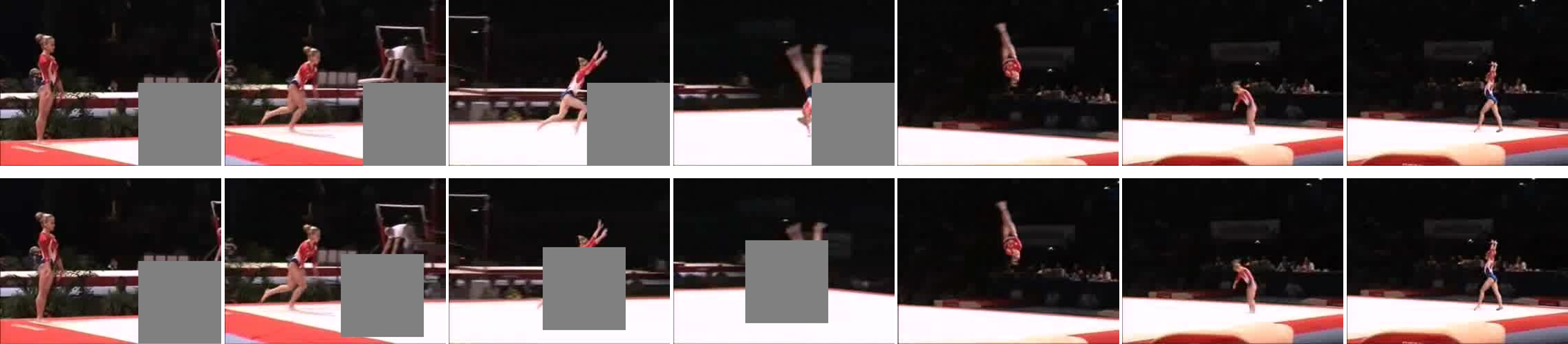}}
\subfigure[Top: CutMix \cite{cutmix,temporal}, Bottom: FloatCutMix]{\includegraphics[width=\textwidth]{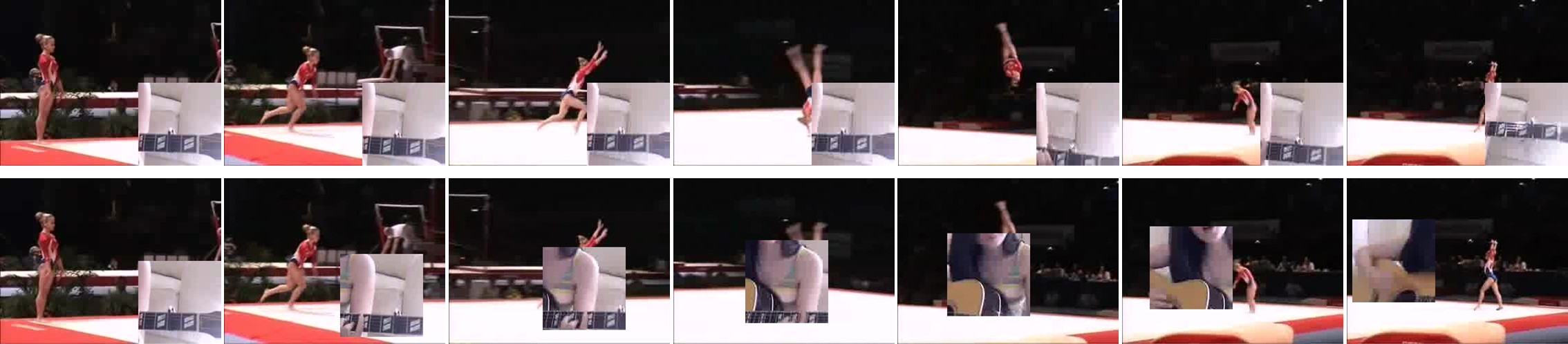}}
\subfigure[Top: CubeCutMix \cite{temporal}, Bottom: FloatCubeCutMix]{\includegraphics[width=\textwidth]{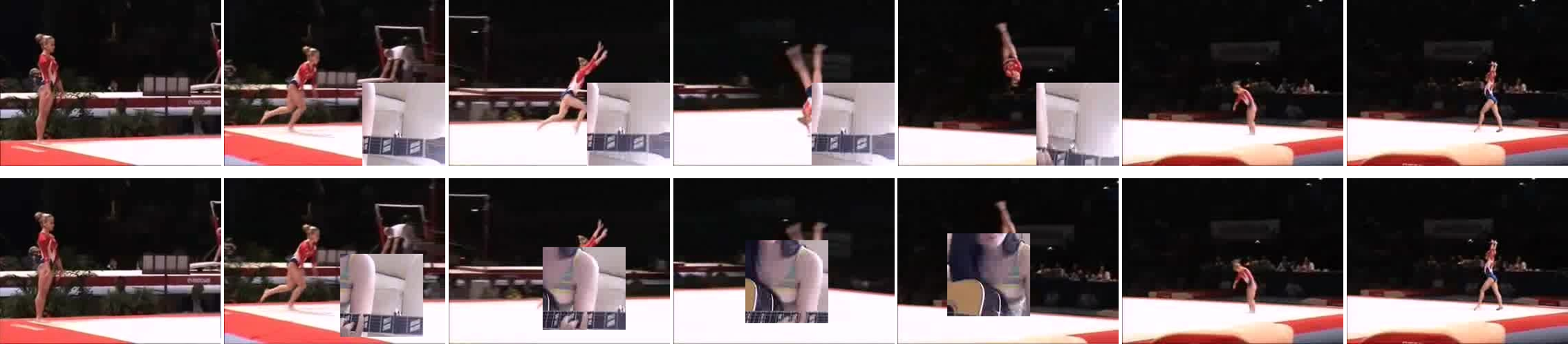}}
\subfigure[Top: CutMixUp \cite{temporal}, Bottom: FloatCutMixUp]{\includegraphics[width=\textwidth]{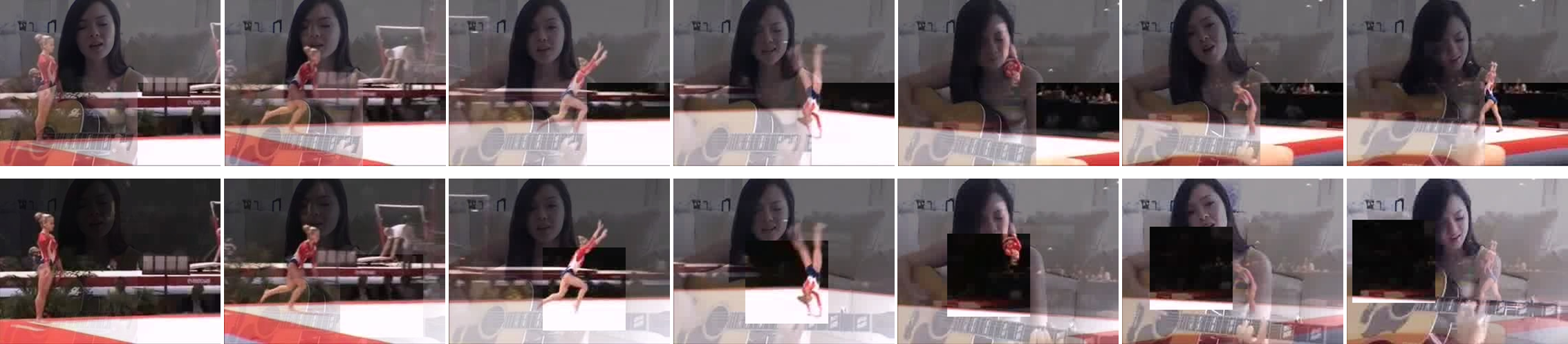}}
\end{figure}
\begin{figure}
\centering
\subfigure[Top: CubeCutMixUp \cite{temporal}, Bottom: FloatCubeCutMixUp]{\includegraphics[width=\textwidth]{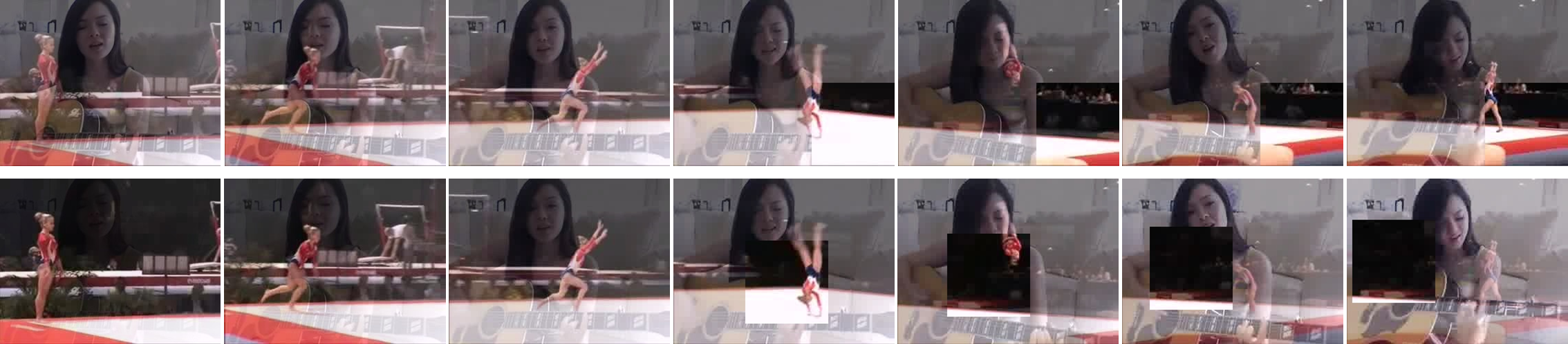}}
\subfigure[Top: FrameCutMixUp \cite{temporal}, Bottom: FloatFrameCutMixUp]{\includegraphics[width=\textwidth]{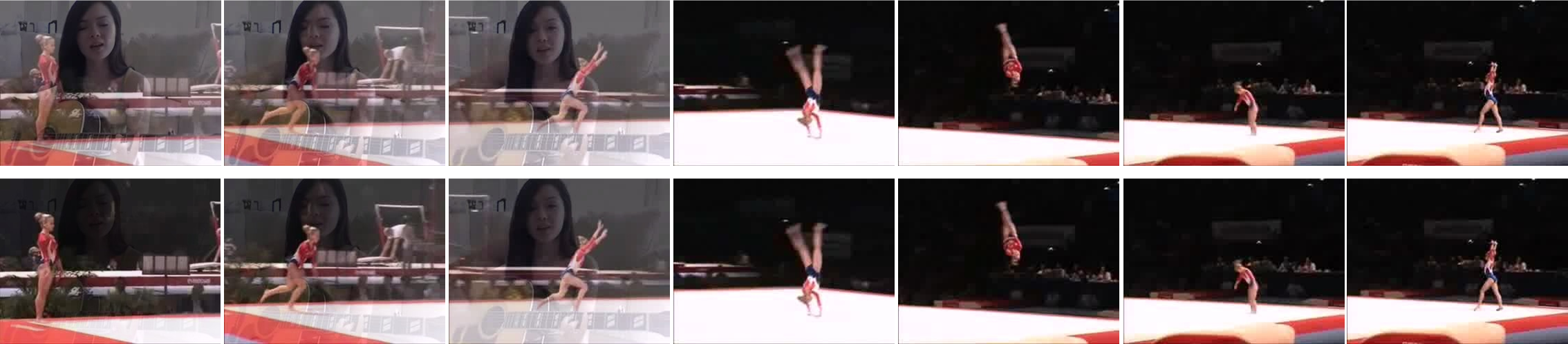}}
\caption{Visual comparison of temporal deleting, cut-and-pasting, and blending algorithms with static and dynamic $r$ and $\lambda$.}
\label{fig:floataugment}
\end{figure}
The temporal adaptations of CutOut \cite{cutout} and CutMix \cite{cutmix} apply a bounding box, $B$, to every frame of a given sample, without changing its position, $r$. In CutMix, the frame sequences are also aligned with the video used for mixing. The concept of static location is practiced in the algorithms' extensions as well - CubeCutOut, CubeCutMix, CutMixUp, and CubeCutMixUp \cite{temporal}. The temporal version of MixUp \cite{mixup} has a fixed mixing ratio, $\lambda$, and remains so in its extensions too - CutMixUp, FrameCutMixUp, CubeCutMixUp \cite{cutmixup,temporal}. Such an idea removes the stochastic behaviour that would be introduced if the aforementioned augmentations were applied to frames separately, without acknowledging them as a part of data series. However, the regularisation techniques themselves can be temporally varied too. By taking a deterministic approach, we are able to enhance the level of spatiotemporal augmentations and involve more bounding box positions and mixing ratios within a batch.

In this subsection, we propose dynamic $r$ and $\lambda$, by linearly changing them across the time dimension. The concept is similar to RandAugment-T, only this time we generate $r_1/r_n$ or $\lambda_1/\lambda_n$ instead of magnitude points for the start and the end of a training sample. Therefore, for delete and cut-and-paste algorithms $r$ becomes:
\begin{equation}
\begin{gathered}
    r_{1_x} \sim U(r_w, W - r_w), r_{n_x} \sim U(r_w, W - r_w), r_w = W\sqrt{1-I},\\
    r_{1_y} \sim U(r_h, H - r_h), r_{n_y} \sim U(r_h, H - r_h), r_h = H\sqrt{1-I},\\
    \centering
    I \sim \text{Beta}(\alpha, \alpha),
\end{gathered}
\end{equation}
where $W, H, U$, and $\alpha$ are the frame width, frame height, uniform distribution, and beta distribution parameter, respectively. Please note that unlike the previous implementations \cite{cutout,cutmix,cutmixup,temporal}, we ensure that the bounding box is fully within the frame at all times, therefore guaranteeing label consistency across the time dimension and omitting the label recalculation step. $r_w$ and $r_h$ are calculated once per video. When start and end points are found, the rest is computed via linear interpolation between the two. Although the authors of FadeMixUp \cite{temporal} introduced dynamic $\lambda$, CutMixUp, FrameCutMixUp, CubeCutMixUp still used the static one. To make the three algorithms temporally varied when it comes to blending, we substitute MixUp with FadeMixUp. All of the above results in seven new regularisation approaches: FloatCutOut, FloatCubeCutOut, FloatCutMix, FloatCubeCutMix, FloatCutMixUp, FloatCubeCutMixUp, and FloatFrameCutMixUp (Fig. \ref{fig:floataugment}).

\section{\uppercase{Experiments}}\label{experiments}
We train and test the approaches mentioned in this paper on the UCF-101 \cite{ucf} and HMDB-51 \cite{hmdb} datasets to assess their effectiveness. The UCF-101 dataset contains 13 320 videos split into 101 categories, whereas HMDB-51 consists of 6 766 videos split into 51 categories. To keep the comparison with the previous work fair \cite{temporal}, we use the same training and testing splits, network architecture \cite{slowfast}, optimiser \cite{adam}, training setup and hyperparameters, and additional techniques, such as learning rate warm-up \cite{lrwarmup}, cosine learning rate scheduling \cite{lrschedule}. Please note that for methods proposed by Kim et al. \cite{temporal}, we report results achieved by running the published code\footnote{\url{https://github.com/taeoh-kim/temporal\_data\_augmentation}} ourselves. For all tables, \textbf{bold text} indicates the highest accuracy. For the UCF-101, the displayed numbers represent the results on the 1\textsuperscript{st} VIPriors action recognition challenge split. For the HMDB-51, we report the average results obtained from 3 different splits \cite{hmdb}.

\subsection{Single video augmentation}
In this subsection, we evaluate VideoReverse, FrameFadeIn, VideoCutMix, and MagAugment, the results can be found in Table \ref{singlevideoresults}. RandAugment indicates static magnitude, applied evenly to all the frames of a given video. For RandAugment-T+, $M1$ and $M_n$ are set to $M-\delta$ and $M+\delta$, respectively, where $\delta=U(0, 0.5*M)$, and $M$ comes from the values used by RandAugment. RandAugment-T++ stands for the extended version, which includes VideoReverse, FrameFadeIn and VideoCutMix (abbreviated as VR, FFI, and VCM, respectively). We also include an ablation study by disabling each of the transformations. For MagAugment, a grid search of $\beta\in[2, 4, 8, 16]$ and the amount of magnitude checkpoints, $P\in[1, 2, 3, 4]$, was used to obtain the highest accuracy, with $\beta=8$ and $P=2$ demonstrating the best performance. We apply MagAugment to all transformations present in RandAugment-T++, apart from the ones that cannot facilitate varying magnitude - Identity, Reverse, AutoContrast, Equalise, ColourInvert, FrameFadeIn, and VideoCutMix. 

The results show that including more single video augmentations provides a benefit with no added computational overhead, thanks to the nature of RandAugment. However, since the improvements in performance are rather small, it is unclear whether all the proposed augmentations are useful. By enabling MagAugment, we obtain 2.36\% and 1.46\% accuracy increases over spatial-only RandAugment, compared to 0.65\% and 0.24\% achieved by RandAugment-T, in UCF-101 Top1 and HMDB-51 Top1 settings, respectively.
\begin{table}
\caption{Video action recognition results on the UCF-101 and HMDB-51 datasets for single video augmentation techniques.}
\begin{center}
\begin{tabular}{|l|c|c|c|c|}
\hline
Method & UCF Top-1 & UCF Top-5 & HMDB Top-1 & HMDB Top-5\\
\hline\hline
Baseline \cite{slowfast} & 54.93 & 77.43 & 39.12 & 69.89 \\
\hline\hline
RandAugment \cite{randaugment,temporal} & 69.82 & 88.57 & 49.24 & 79.94 \\
\hline
RandAugment-T+ \cite{temporal} & 70.47 & 89.94 & 49.48 & 80.17 \\
\hline
RandAugment-T++ & \textbf{70.74} & \textbf{90.04} & \textbf{49.60} & \textbf{80.21} \\
\hline\hline
RandAugment-T++ - VR & 70.52 & 89.94 & 49.50 & 80.10 \\
\hline
RandAugment-T++ - FFI & 70.58 & 89.98 & 49.58 & 80.06 \\
\hline
RandAugment-T++ - VCM & 70.76 & 90.12 & 49.54 & 80.17 \\
\hline\hline
MagAugment & \textbf{72.18} & \textbf{93.78} & \textbf{50.70} & \textbf{81.12} \\
\hline
\end{tabular}
\end{center}
\label{singlevideoresults}
\end{table}

\begin{table}
\caption{Video action recognition results on the UCF-101 and HMDB-51 datasets for temporal deleting, cut-and-pasting, and blending techniques.}
\begin{center}
\begin{tabular}{|l|c|c|c|c|}
\hline
Method & UCF Top-1 & UCF Top-5 & HMDB Top-1 & HMDB Top-5\\
\hline\hline
Baseline \cite{slowfast} & 54.93 & 77.43 & 39.12 & 69.89 \\
\hline\hline
CutOut \cite{cutout,temporal} & 51.16 & 74.25 & 36.93 & 68.07 \\
\hline
CubeCutOut \cite{temporal} & 51.82 & 76.73 & 37.50 & 68.53 \\
\hline
\textit{F}CutOut & 54.24 & 76.23 & 39.02 & 69.89 \\
\hline
\textit{F}CubeCutOut & \textbf{54.68} & \textbf{77.19} & \textbf{39.25} & \textbf{70.00} \\
\hline\hline
CutMix \cite{cutmix,temporal} & 53.03 & 76.78 & 34.69 & 65.67 \\
\hline
CubeCutMix \cite{temporal} & 54.91 & 77.34 & 36.75 & 67.23 \\
\hline
\textit{F}CutMix & 55.25 & 77.27 & 37.32 & 68.24 \\
\hline
\textit{F}CubeCutMix & \textbf{55.66} & \textbf{78.00} & \textbf{39.42} & \textbf{69.98} \\
\hline\hline
CutMixUp \cite{cutmixup,temporal} & 60.08 & 82.14 & 43.13 & 74.19 \\
\hline
CubeCutMixUp \cite{temporal} & 60.16 & 82.14 & 43.15 & 74.24 \\
\hline
FrameCutMixUp \cite{temporal} & 61.02 & 82.97 & 42.88 & 74.08 \\
\hline
\textit{F}CutMixUp & 62.41 & 84.59 & 45.06 & 75.78 \\
\hline
\textit{F}CubeCutMixUp & 62.38 & 84.70 & 45.12 & 75.85 \\
\hline
\textit{F}FrameCutMixUp & \textbf{63.04} & \textbf{85.64} & \textbf{45.98} & \textbf{76.90} \\
\hline
\end{tabular}
\end{center}
\label{temporalresults}
\end{table}

\subsection{Temporal deleting, cut-and-pasting, and blending}
We present the results of Cutout \cite{cutout}, CutMix \cite{cutmix}, and CutMixUp \cite{cutmixup}, and their temporal extensions, which can be found in Table \ref{temporalresults}. We prefix our methods with \textit{F} to save space and indicate floating bounding box positions and mixing ratios. Single video augmentation is turned off in this experiment. Although the CutOut variants struggle to beat the baseline and the CutMix spin-offs demonstrate a rather small boost in accuracy, it is clear that having dynamic $r$ and $\lambda$ helps the model to consistently achieve better performance - when compared side by side, the floating extensions demonstrate an average gain of 2.45\%, when compared to their static variants in the Top-1 settings. FloatFrameCutMixUp scores the highest accuracy, improving over the baseline by 8.11\% and 6.86\% in UCF-101 Top-1 and HMDB-51 Top-1 settings, respectively. FloatCubeCutOut, FloatCubeCutMix, and FloatFrameCutMixUp perform the best in their respective groups, suggesting that retaining some of the frames of a video unaffected might yield additional benefits (Fig. \ref{fig:floataugment}b, Fig. \ref{fig:floataugment}d, Fig. \ref{fig:floataugment}g). 

\section{\uppercase{Conclusions}}\label{conclusions}
In this paper, we introduced several novel temporal data augmentation methods. We showed that developing video-specific transformations and including more aggressive magnitude transitions is beneficial for networks that aim to solve video action recognition. We extended temporal versions of CutOut, CutMix, and CutMixUp further by changing their nature from static to dynamic, and observed an improvement performance. Future work includes combining single video augmentations with delete, cut-and-paste, and blend techniques to expand the total amount of possible augmentation combinations, covering more baseline models to analyse applicability and versatility of the proposed methods, and testing the framework on larger datasets, such as Kinetics \cite{kinetics} and Something-Something-v2 \cite{something}.

\subsubsection*{Acknowledgements}
The authors are grateful for the support from the Natural Environment Research Council and Engineering and Physical Sciences Research Council through the NEXUSS Centre for Doctoral Training (grant \linebreak \#NE/RO12156/1).
%
%

\bibliographystyle{splncs04}
\bibliography{main}

\begin{thebibliography}{10}
\providecommand{\url}[1]{\texttt{#1}}
\providecommand{\urlprefix}{URL }
\providecommand{\doi}[1]{https://doi.org/#1}

\bibitem{dagan}
Antoniou, A., Storkey, A., Edwards, H.: Data augmentation generative
  adversarial networks. arXiv preprint arXiv:1711.04340  (2017)

\bibitem{largegan}
Brock, A., Donahue, J., Simonyan, K.: Large scale gan training for high
  fidelity natural image synthesis. arXiv preprint arXiv:1809.11096  (2018)

\bibitem{kinetics}
Carreira, J., Zisserman, A.: Quo vadis, action recognition? a new model and the
  kinetics dataset. In: proceedings of the IEEE Conference on Computer Vision
  and Pattern Recognition. pp. 6299--6308 (2017)

\bibitem{contr}
Chen, T., Kornblith, S., Norouzi, M., Hinton, G.: A simple framework for
  contrastive learning of visual representations. In: International conference
  on machine learning. pp. 1597--1607. PMLR (2020)

\bibitem{feataugment}
Chu, P., Bian, X., Liu, S., Ling, H.: Feature space augmentation for
  long-tailed data. In: Vedaldi, A., Bischof, H., Brox, T., Frahm, J.M. (eds.)
  Computer Vision -- ECCV 2020. pp. 694--710. Springer International
  Publishing, Cham (2020)

\bibitem{visualobjcls}
Cireşan, D., Meier, U., Masci, J., Gambardella, L.M., Schmidhuber, J.:
  High-performance neural networks for visual object classification. Computing
  Research Repository - CORR  (02 2011)

\bibitem{multicolumn}
Cireşan, D., Meier, U., Schmidhuber, J.: Multi-column deep neural networks for
  image classification. Proceedings / CVPR, IEEE Computer Society Conference on
  Computer Vision and Pattern Recognition. IEEE Computer Society Conference on
  Computer Vision and Pattern Recognition  (02 2012).
  \doi{10.1109/CVPR.2012.6248110}

\bibitem{autoaugment}
Cubuk, E.D., Zoph, B., Mane, D., Vasudevan, V., Le, Q.V.: Autoaugment: Learning
  augmentation policies from data. arXiv preprint arXiv:1805.09501  (2018)

\bibitem{randaugment}
Cubuk, E.D., Zoph, B., Shlens, J., Le, Q.V.: Randaugment: Practical automated
  data augmentation with a reduced search space. In: Proceedings of the
  IEEE/CVF conference on computer vision and pattern recognition workshops. pp.
  702--703 (2020)

\bibitem{datasetaugment}
DeVries, T., Taylor, G.W.: Dataset augmentation in feature space. arXiv
  preprint arXiv:1702.05538  (2017)

\bibitem{cutout}
DeVries, T., Taylor, G.W.: Improved regularization of convolutional neural
  networks with cutout. arXiv preprint arXiv:1708.04552  (2017)

\bibitem{autoencoders}
Doersch, C.: Tutorial on variational autoencoders. arXiv preprint
  arXiv:1606.05908  (2016)

\bibitem{slowfast}
Feichtenhofer, C., Fan, H., Malik, J., He, K.: Slowfast networks for video
  recognition. In: Proceedings of the IEEE/CVF international conference on
  computer vision. pp. 6202--6211 (2019)

\bibitem{twostreamfusion}
Feichtenhofer, C., Pinz, A., Zisserman, A.: Convolutional two-stream network
  fusion for video action recognition. In: Proceedings of the IEEE conference
  on computer vision and pattern recognition. pp. 1933--1941 (2016)

\bibitem{se}
French, G., Mackiewicz, M., Fisher, M.: Self-ensembling for visual domain
  adaptation. In: International Conference on Learning Representations (2018)

\bibitem{cowmix}
French, G., Oliver, A., Salimans, T.: Milking cowmask for semi-supervised image
  classification. arXiv preprint arXiv:2003.12022  (2020)

\bibitem{medgan}
Frid-Adar, M., Diamant, I., Klang, E., Amitai, M., Goldberger, J., Greenspan,
  H.: Gan-based synthetic medical image augmentation for increased cnn
  performance in liver lesion classification. Neurocomputing  \textbf{321},
  321--331 (2018)

\bibitem{dann}
Ganin, Y., Lempitsky, V.: Unsupervised domain adaptation by backpropagation.
  In: International conference on machine learning. pp. 1180--1189. PMLR (2015)

\bibitem{gan}
Goodfellow, I., Pouget-Abadie, J., Mirza, M., Xu, B., Warde-Farley, D., Ozair,
  S., Courville, A., Bengio, Y.: Generative adversarial nets. Advances in
  neural information processing systems  \textbf{27} (2014)

\bibitem{adversarialexamples}
Goodfellow, I.J., Shlens, J., Szegedy, C.: Explaining and harnessing
  adversarial examples. arXiv preprint arXiv:1412.6572  (2014)

\bibitem{jellygan}
Gorpincenko, A., French, G., Knight, P., Challiss, M., Mackiewicz, M.:
  Improving automated sonar video analysis to notify about jellyfish blooms.
  IEEE Sensors Journal  \textbf{21}(4),  4981--4988 (2021).
  \doi{10.1109/JSEN.2020.3032031}

\bibitem{vatta3n}
Gorpincenko, A., French, G., Mackiewicz, M.: Virtual adversarial training in
  feature space to improve unsupervised video domain adaptation (2020)

\bibitem{lrwarmup}
Goyal, P., Doll{\'a}r, P., Girshick, R., Noordhuis, P., Wesolowski, L., Kyrola,
  A., Tulloch, A., Jia, Y., He, K.: Accurate, large minibatch sgd: Training
  imagenet in 1 hour. arXiv preprint arXiv:1706.02677  (2017)

\bibitem{something}
Goyal, R., Ebrahimi~Kahou, S., Michalski, V., Materzynska, J., Westphal, S.,
  Kim, H., Haenel, V., Fruend, I., Yianilos, P., Mueller-Freitag, M., et~al.:
  The" something something" video database for learning and evaluating visual
  common sense. In: Proceedings of the IEEE international conference on
  computer vision. pp. 5842--5850 (2017)

\bibitem{momentum}
He, K., Fan, H., Wu, Y., Xie, S., Girshick, R.: Momentum contrast for
  unsupervised visual representation learning. In: Proceedings of the IEEE/CVF
  conference on computer vision and pattern recognition. pp. 9729--9738 (2020)

\bibitem{population}
Ho, D., Liang, E., Chen, X., Stoica, I., Abbeel, P.: Population based
  augmentation: Efficient learning of augmentation policy schedules. In:
  International Conference on Machine Learning. pp. 2731--2741. PMLR (2019)

\bibitem{pairing}
Inoue, H.: Data augmentation by pairing samples for images classification.
  arXiv preprint arXiv:1801.02929  (2018)

\bibitem{3dcnn}
Ji, S., Xu, W., Yang, M., Yu, K.: 3d convolutional neural networks for human
  action recognition. IEEE Transactions on Pattern Analysis and Machine
  Intelligence  \textbf{35}(1),  221--231 (2013). \doi{10.1109/TPAMI.2012.59}

\bibitem{slowfusion}
Karpathy, A., Toderici, G., Shetty, S., Leung, T., Sukthankar, R., Fei-Fei, L.:
  Large-scale video classification with convolutional neural networks. In: 2014
  IEEE Conference on Computer Vision and Pattern Recognition. pp. 1725--1732
  (2014). \doi{10.1109/CVPR.2014.223}

\bibitem{stylegan}
Karras, T., Laine, S., Aila, T.: A style-based generator architecture for
  generative adversarial networks. In: Proceedings of the IEEE/CVF conference
  on computer vision and pattern recognition. pp. 4401--4410 (2019)

\bibitem{betterstylegan}
Karras, T., Laine, S., Aittala, M., Hellsten, J., Lehtinen, J., Aila, T.:
  Analyzing and improving the image quality of stylegan. In: Proceedings of the
  IEEE/CVF conference on computer vision and pattern recognition. pp.
  8110--8119 (2020)

\bibitem{spatiotemporal}
Kim, J.Y., Ha, J.E.: Spatio-temporal data augmentation for visual surveillance.
  IEEE Access  \textbf{PP}, ~1--1 (12 2021). \doi{10.1109/ACCESS.2021.3135505}

\bibitem{rms}
Kim, J., Cha, S., Wee, D., Bae, S., Kim, J.: Regularization on
  spatio-temporally smoothed feature for action recognition. In: Proceedings of
  the IEEE/CVF conference on computer vision and pattern recognition. pp.
  12103--12112 (2020)

\bibitem{temporal}
Kim, T., Lee, H., Cho, M., Lee, H.S., Cho, D.H., Lee, S.: Learning temporally
  invariant and localizable features via data augmentation for video
  recognition. In: European Conference on Computer Vision. pp. 386--403.
  Springer (2020)

\bibitem{adam}
Kingma, D.P., Ba, J.: Adam: A method for stochastic optimization. arXiv
  preprint arXiv:1412.6980  (2014)

\bibitem{imagenet}
Krizhevsky, A., Sutskever, I., Hinton, G.E.: Imagenet classification with deep
  convolutional neural networks. In: Pereira, F., Burges, C., Bottou, L.,
  Weinberger, K. (eds.) Advances in Neural Information Processing Systems.
  vol.~25. Curran Associates, Inc. (2012)

\bibitem{simpleweightdecay}
Krogh, A., Hertz, J.: A simple weight decay can improve generalization.
  Advances in neural information processing systems  \textbf{4} (1991)

\bibitem{hmdb}
Kuehne, H., Jhuang, H., Garrote, E., Poggio, T., Serre, T.: Hmdb: A large video
  database for human motion recognition. In: 2011 International Conference on
  Computer Vision. pp. 2556--2563 (2011). \doi{10.1109/ICCV.2011.6126543}

\bibitem{sonarsecond}
Lee, S., Park, B., Kim, A.: Deep learning based object detection via
  style-transferred underwater sonar images ⁎⁎this work is supported
  through a grant from msip (no 2015r1c1a2a01052138), iitp grant funded by msit
  (no.2017-0-00067), and a grant from endowment project of kriso (pes9390).
  authors are grateful to sonartech for sharing sample videos for the research.
  IFAC-PapersOnLine  \textbf{52}(21),  152--155 (2019).
  \doi{https://doi.org/10.1016/j.ifacol.2019.12.299}, 12th IFAC Conference on
  Control Applications in Marine Systems, Robotics, and Vehicles CAMS 2019

\bibitem{smartaugment}
Lemley, J., Bazrafkan, S., Corcoran, P.M.: Smart augmentation learning an
  optimal data augmentation strategy. IEEE Access  \textbf{5},  5858--5869
  (2017)

\bibitem{fastautoaugment}
Lim, S., Kim, I., Kim, T., Kim, C., Kim, S.: Fast autoaugment. Advances in
  Neural Information Processing Systems  \textbf{32} (2019)

\bibitem{feattransfer}
Liu, B., Wang, X., Dixit, M., Kwitt, R., Vasconcelos, N.: Feature space
  transfer for data augmentation. In: Proceedings of the IEEE Conference on
  Computer Vision and Pattern Recognition (CVPR) (June 2018)

\bibitem{lrschedule}
Loshchilov, I., Hutter, F.: Sgdr: Stochastic gradient descent with warm
  restarts. arXiv preprint arXiv:1608.03983  (2016)

\bibitem{weightdecay}
Loshchilov, I., Hutter, F.: Decoupled weight decay regularization. arXiv
  preprint arXiv:1711.05101  (2017)

\bibitem{vmt}
Mao, X., Ma, Y., Yang, Z., Chen, Y., Li, Q.: Virtual mixup training for
  unsupervised domain adaptation (2019)

\bibitem{self}
Misra, I., Maaten, L.v.d.: Self-supervised learning of pretext-invariant
  representations. In: Proceedings of the IEEE/CVF Conference on Computer
  Vision and Pattern Recognition. pp. 6707--6717 (2020)

\bibitem{vat}
{Miyato}, T., {Maeda}, S., {Koyama}, M., {Ishii}, S.: Virtual adversarial
  training: A regularization method for supervised and semi-supervised
  learning. IEEE Transactions on Pattern Analysis and Machine Intelligence
  \textbf{41}(8),  1979--1993 (2019). \doi{10.1109/TPAMI.2018.2858821}

\bibitem{deepfool}
Moosavi-Dezfooli, S.M., Fawzi, A., Frossard, P.: Deepfool: a simple and
  accurate method to fool deep neural networks. In: Proceedings of the IEEE
  conference on computer vision and pattern recognition. pp. 2574--2582 (2016)

\bibitem{noise}
Moreno-Barea, F.J., Strazzera, F., Jerez, J.M., Urda, D., Franco, L.: Forward
  noise adjustment scheme for data augmentation. In: 2018 IEEE Symposium Series
  on Computational Intelligence (SSCI). pp. 728--734 (2018).
  \doi{10.1109/SSCI.2018.8628917}

\bibitem{earlystopping}
Prechelt, L.: Early stopping-but when? In: Neural Networks: Tricks of the
  trade, pp. 55--69. Springer (1998)

\bibitem{dirt}
Shu, R., Bui, H.H., Narui, H., Ermon, S.: A dirt-t approach to unsupervised
  domain adaptation. arXiv preprint arXiv:1802.08735  (2018)

\bibitem{visualcontana}
Simard, P., Steinkraus, D., Platt, J.: Best practices for convolutional neural
  networks applied to visual document analysis. In: Seventh International
  Conference on Document Analysis and Recognition, 2003. Proceedings. pp.
  958--963 (2003). \doi{10.1109/ICDAR.2003.1227801}

\bibitem{twostream}
Simonyan, K., Zisserman, A.: Two-stream convolutional networks for action
  recognition in videos. In: Proceedings of the 27th International Conference
  on Neural Information Processing Systems - Volume 1. p. 568–576. NIPS'14,
  MIT Press, Cambridge, MA, USA (2014)

\bibitem{fixmatch}
Sohn, K., Berthelot, D., Carlini, N., Zhang, Z., Zhang, H., Raffel, C.A.,
  Cubuk, E.D., Kurakin, A., Li, C.L.: Fixmatch: Simplifying semi-supervised
  learning with consistency and confidence. Advances in neural information
  processing systems  \textbf{33},  596--608 (2020)

\bibitem{ucf}
Soomro, K., Zamir, A.R., Shah, M.: Ucf101: A dataset of 101 human actions
  classes from videos in the wild. arXiv preprint arXiv:1212.0402  (2012)

\bibitem{dropout}
Srivastava, N., Hinton, G., Krizhevsky, A., Sutskever, I., Salakhutdinov, R.:
  Dropout: A simple way to prevent neural networks from overfitting. Journal of
  Machine Learning Research  \textbf{15}(56),  1929--1958 (2014)

\bibitem{improvedmix}
Summers, C., Dinneen, M.J.: Improved mixed-example data augmentation. In: 2019
  IEEE Winter Conference on Applications of Computer Vision (WACV). pp.
  1262--1270. IEEE (2019)

\bibitem{2d1d}
Sun, L., Jia, K., Yeung, D., Shi, B.E.: Human action recognition using
  factorized spatio-temporal convolutional networks. In: 2015 IEEE
  International Conference on Computer Vision (ICCV). pp. 4597--4605. IEEE
  Computer Society, Los Alamitos, CA, USA (dec 2015).
  \doi{10.1109/ICCV.2015.522}

\bibitem{sonarfirst}
Terayama, K., Shin, K., Mizuno, K., Tsuda, K.: Integration of sonar and optical
  camera images using deep neural network for fish monitoring. Aquacultural
  Engineering  \textbf{86},  102000 (2019).
  \doi{https://doi.org/10.1016/j.aquaeng.2019.102000}

\bibitem{bayesian}
Tran, T., Pham, T., Carneiro, G., Palmer, L., Reid, I.: A bayesian data
  augmentation approach for learning deep models. In: Proceedings of the 31st
  International Conference on Neural Information Processing Systems. p.
  2794–2803. NIPS'17, Curran Associates Inc., Red Hook, NY, USA (2017)

\bibitem{adda}
Tzeng, E., Hoffman, J., Saenko, K., Darrell, T.: Adversarial discriminative
  domain adaptation. In: Proceedings of the IEEE conference on computer vision
  and pattern recognition. pp. 7167--7176 (2017)

\bibitem{longfusion}
Varol, G., Laptev, I., Schmid, C.: Long-term temporal convolutions for action
  recognition. IEEE transactions on pattern analysis and machine intelligence
  \textbf{40}(6),  1510--1517 (2017)

\bibitem{lowgan}
Wang, Y.X., Girshick, R., Hebert, M., Hariharan, B.: Low-shot learning from
  imaginary data. In: Proceedings of the IEEE conference on computer vision and
  pattern recognition. pp. 7278--7286 (2018)

\bibitem{cutmixup}
Yoo, J., Ahn, N., Sohn, K.A.: Rethinking data augmentation for image
  super-resolution: A comprehensive analysis and a new strategy. In:
  Proceedings of the IEEE/CVF Conference on Computer Vision and Pattern
  Recognition. pp. 8375--8384 (2020)

\bibitem{cutmix}
{Yun}, S., {Han}, D., {Chun}, S., {Oh}, S.J., {Yoo}, Y., {Choe}, J.: Cutmix:
  Regularization strategy to train strong classifiers with localizable
  features. 2019 IEEE/CVF International Conference on Computer Vision (ICCV)
  pp. 6022--6031 (2019). \doi{10.1109/ICCV.2019.00612}

\bibitem{mixup}
Zhang, H., Cisse, M., Dauphin, Y.N., Lopez-Paz, D.: mixup: Beyond empirical
  risk minimization. arXiv preprint arXiv:1710.09412  (2017)

\bibitem{cyclegan}
Zhu, J.Y., Park, T., Isola, P., Efros, A.A.: Unpaired image-to-image
  translation using cycle-consistent adversarial networks. In: Proceedings of
  the IEEE international conference on computer vision. pp. 2223--2232 (2017)

\end{thebibliography}





\end{document}